\documentclass[letterpaper, 10 pt, conference]{ieeeconf}  

\IEEEoverridecommandlockouts                              

\usepackage[ruled,vlined]{algorithm2e}
\let\labelindent\relax
\usepackage{enumitem}
\usepackage{amsmath}
\usepackage{xcolor}
\usepackage{booktabs}
\usepackage{graphicx}
\usepackage{amssymb}
\usepackage{array}
\usepackage{subcaption}
\usepackage{lipsum} 
\usepackage{makecell}
\usepackage{svg}
\usepackage{multirow}
\usepackage{hyperref}
\usepackage{epigraph}

\title{\LARGE \bf
Targeted Attack on Deep RL-based Autonomous Driving \\ with Learned Visual Patterns
}

\author{
Prasanth Buddareddygari$^{1}$, Travis Zhang$^{2}$, Yezhou Yang$^{1}$, and Yi Ren$^{3}$
\thanks{$^{1}$ PB and YY are with the Active Perception Group at the School of Computing and Augmented Intelligence, Arizona State University, Tempe, AZ 85287, USA
        {\tt\small \{pbuddare, yz.yang\}@asu.edu}}%
\thanks{$^{2}$ TZ is with the College of Engineering, Cornell University, Ithaca, NY, 14853, USA. This work was done while TZ was an intern at the Active Perception Group, Arizona State University.
        {\tt\small tz98@cornell.edu}}%

\thanks{$^{3}$ YR is with the School for Engineering of Matter, Transport, and Energy, Arizona State University, Tempe, AZ 85287, USA
        {\tt\small yiren@asu.edu}}%
}

\begin{document}

\maketitle
\thispagestyle{empty}
\pagestyle{empty}

\begin{abstract}
Recent studies demonstrated the vulnerability of control policies learned through deep reinforcement learning against adversarial attacks, raising concerns about the application of such models to risk-sensitive tasks such as autonomous driving. Threat models for these demonstrations are limited to (1) targeted attacks through real-time manipulation of the agent's observation, and (2) untargeted attacks through manipulation of the physical environment. The former assumes full access to the agent's states/observations at all times, while the latter has no control over attack outcomes.
This paper investigates the feasibility of targeted attacks through visually learned patterns placed on physical objects in the environment, a threat model that combines the practicality and effectiveness of the existing ones. Through analysis, we demonstrate  that a pre-trained policy can be hijacked within a time window, e.g., performing an unintended self-parking, when an adversarial object is present. To enable the attack, we adopt an assumption that the dynamics of both the environment and the agent can be learned by the attacker.
Lastly, we empirically show the effectiveness of the proposed attack on different driving scenarios, perform a location robustness test, and study the tradeoff between the attack strength and its effectiveness.
Code is available at \textcolor{blue}{\url{https://github.com/ASU-APG/Targeted-Physical-Adversarial-Attacks-on-AD}}
\end{abstract}

\section{Introduction}

\setlength{\epigraphwidth}{.95\columnwidth}
\renewcommand{\epigraphflush}{center}
\renewcommand{\textflush}{flushepinormal}
\epigraph{{\normalsize \textit{``Attack is the secret of defense; defense is the planning of an attack.''} }}
{{\footnotesize{\textit{-- Sun Tzu, The Art of War, 5th century BC}}}}

Deep reinforcement learning (RL) has grown tremendously in the past few years, producing close-to-human control policies on various tasks~\cite{pmlr-v119-laskin20a, pmlr-v119-badia20a, schrittwieser2020mastering, pmlr-v48-duan16} including solving Atari games~\cite{pmlr-v119-badia20a, mnih2015human}, robot manipulation~\cite{gu2017deep}, autonomous driving~\cite{wang2018deep, Liang_2018_ECCV}, and many others~\cite{schrittwieser2020mastering}. However, deep neural networks (DNNs) are vulnerable to adversarial attacks, with demonstrations in real world applications such as computer vision~\cite{goodfellow2014explaining, Nguyen_2015_CVPR, akhtar2018threat, su2019one}, natural language processing~\cite{jia-liang-2017-adversarial}, and speech~\cite{pmlr-v97-qin19a}. Recent studies showed that deep RL agents, due to their adoption of DNNs for value or policy approximation, are also susceptible to such attacks~\cite{huang2017adversarial, lin2017tactics, huang2018autoadversarial}.

\setlength{\textfloatsep}{6pt plus 1.0pt minus 2.0pt} 
\begin{figure}[ht!] 
\includegraphics[width=24em,keepaspectratio]{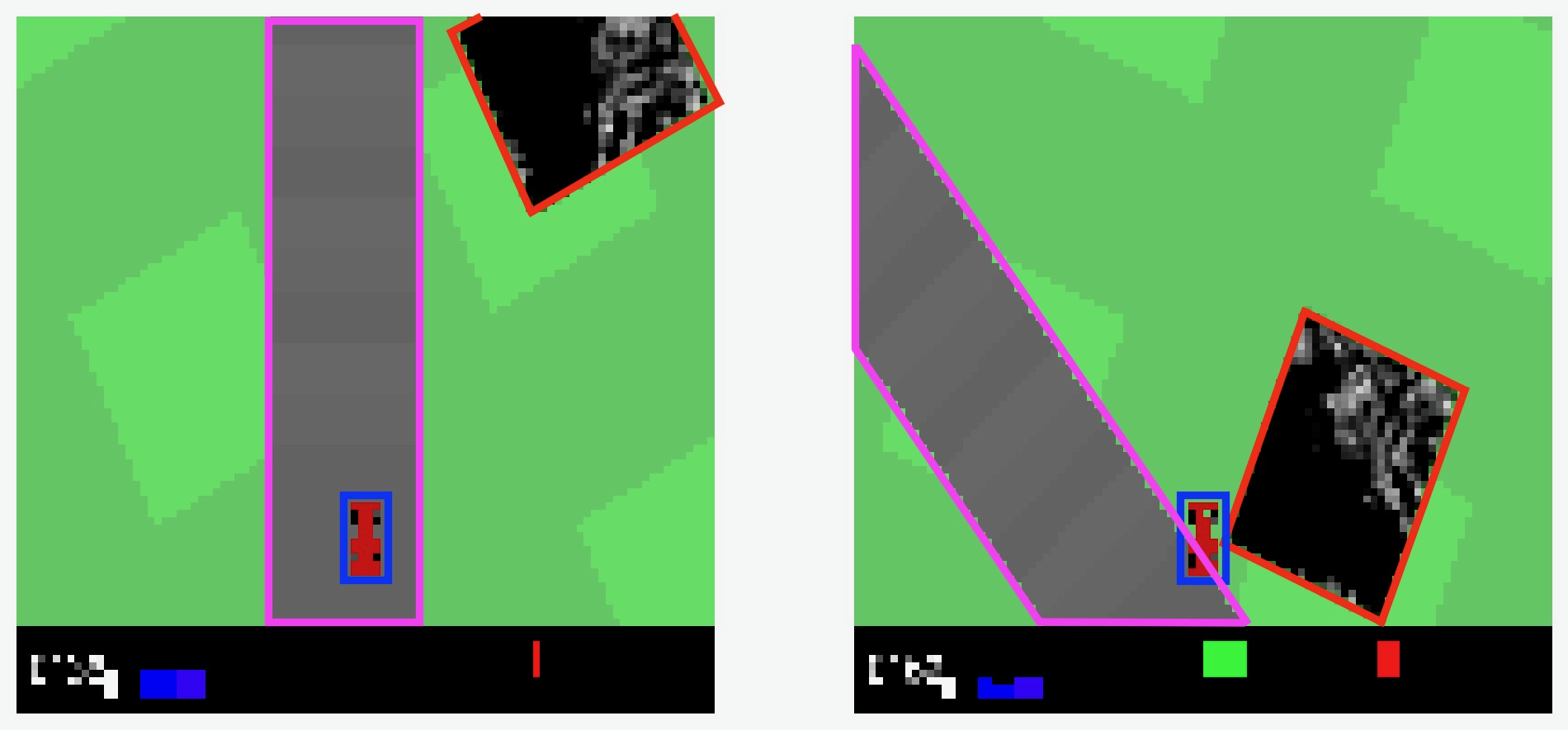}
\caption{Targeted adversarial attack on autonomous driving agent using an object fixed to the ground. The attack formulation incorporates the dynamics of the object subject to the pretrained policy of the agent and the object itself. \textbf{Left:} initial state. \textbf{Right:} achieved target state. The red, blue, and pink bounding boxes indicate the learned adversarial visual patterns, car, and road track, respectively.}
\label{fig:intro}
\end{figure}

The threat models in the deep RL domain form two categories: the first assumes that the attacker can directly manipulate the states/observations or actions of the agent, while the second performs the attack through physical objects placed in the environment. 
Among the first category, Huang et al.~\cite{huang2017adversarial} proposed to directly perturb the agent's observations at \textit{all} time steps during a roll-out. Similarly, Lin et al.~\cite{lin2017tactics} proposed to attack during a chosen subset of time steps. Applications of this type of attacks to autonomous driving have been shown to be effective~\cite{huang2018autoadversarial, sun_stealthy_2020}.
Weng et al.~\cite{weng2020toward} showed that learning the dynamics of agents and environments improves the efficacy of the attack in comparison to model-free methods.
This category of threats, however, are not practical as they require {\bf direct access to the agents' perception modules to modify their observations}. Such a strong prerequisite condition to launch attacks significantly limits the power of these threats. 

For a more feasible approach, researchers studied attacks where adversarial objects are placed in the environments to fool DNNs. Such attacks have been proven effective in general applications such as image classification~\cite{kurakin2016adversarial, pmlr-v80-athalye18b, Eykholt_2018_CVPR} and face recognition~\cite{sharif2016accessorize}. Specific to deep RL, Kong et al.~\cite{Kong_2020_CVPR} and Yang et al.~\cite{yang2020finding} demonstrated the existence of physical adversaries, in the form of
advertising sign boards and patterns painted on the road respectively,
that can successfully mislead autonomous driving systems. 
While their models are more practical, most of the existing attempts of this type are \textbf{not targeted} towards reaching a certain goal state. Instead, they seek to maximize the deviation of actions in the presence of adversaries from the benign policy. These loose-end attacks would only be considered effective when the final state turns out to be disastrous. This is not guaranteed and thus the attack results vary.  



Launching targeted attacks without direct access to the agent's perception modules remains an open challenge. To achieve this, we use the assumptions that the attacker can learn differentiable dynamical models that predict the transition of the environment and has access to the dynamics of the agent's own states with respect to the agent's actions. We argue that these assumptions are reasonable since the environment (e.g., a particular segment of the highway) is accessible to all, including the attacker, and agent dynamics (e.g., for vehicles) is common knowledge. Lastly, since we focus on the existence of policy vulnerability, we assume the agent's policy model to be white-box.

To the best of our knowledge, our work presented in this paper is the first to investigate the existence of \textit{targeted} attacks on deep RL models using adversarial objects in the environment. Specifically, we examine the existence of static and structured perturbations on the environment so that within a time window, the agent is led to a state that is specified by the attacker (Fig.~\ref{fig:intro}). If successful, the study will expose real-world threat models. For instance, a hacker could place an adversarial billboard sign next to the road to cause self-driving cars to veer off the track without directly modifying the cars' observations.

The contributions of our paper are as follows:
\begin{itemize}
	\item 
	Our presented attack algorithm generates a \textbf{static perturbation} that can be directly realized through an object placed in the environment to mislead the agent towards a pre-specified state within a time window.
	
	\item We perform ablation studies to show that the choices of the time window and the attack target are correlated. Therefore, fine-tuning of the loss function of the attack with respect to the time window is necessary for identifying successful attacks.
	
	\item We study the \textbf{robustness of the derived attacks} with respect to their relative locations to the agent, and show that moving the attack object partially out of the sight of the agent will reduce the effect of the attack.  
	
	
\end{itemize}



\vspace{-0.05in}
\section{Related Work}
\vspace{-0.05in}
\subsection{Adversarial attacks on RL agents}
\vspace{-0.03in}
Adversarial attacks in RL, especially in the deep RL domain, have gained attention~\cite{huang2017adversarial, lin2017tactics, huang2018autoadversarial} following those for DNNs~\cite{goodfellow2014explaining, carlini2017towards, su2019one}. In the Atari environment, Lin et al.~\cite{lin2017tactics} proposed a strategically timed attack which focuses on finding the right time when an adversarial attack needs to be performed, and an enchanting attack, a targeted attack that generates adversarial examples in order to find actions that lead to a target state. Kos et al.~\cite{kos2017delving} proposed methods for generating value function-based adversarial examples and Behzadan et al.~\cite{behzadan2017vulnerability} studied adversarial attacks on deep Q networks (DQN) along with transferability to different DQN models. Further, Pattanaik et al.~\cite{pattanaik2017robust} proposed a gradient-based attack on double deep Q networks (DDQN) and deep deterministic policy gradient (DDPG), and developed a robust control framework through adversarial training.
Weng et al.~\cite{weng2020toward} proposed model-based adversarial attacks on MuJoCo~\cite{todorov2012mujoco} domains using a target state as the attack goal similar to the enchanting attack presented by Lin et al.~\cite{lin2017tactics}. More recently, Zhang et al.~\cite{zhang2020robust} proposed a state-adversarial Markov decision process and studied adversarial attacks on model-free deep RL algorithms. While all these aforementioned works have shown that deep RL systems are vulnerable to adversarial attacks, few have explored a target-controlled attack using a dynamical model as presented in this work. 

\vspace{-0.05in}
\subsection{Physical Adversarial Attacks}
\vspace{-0.03in}
There are a few recent works that focused on physical adversarial attacks~\cite{pmlr-v80-athalye18b, sharif2016accessorize}.
With respect to multi-agent environments, Gleave et al.~\cite{gleave2020advpolicies} primarily focused on training an adversarial agent to exploit the weaknesses of traditionally-trained deep RL agents. However, their study, being in a multi-agent setting, does not allow for physical objects to be placed in the environment and is different from the threat model proposed in this paper. 
Kong et al.~\cite{Kong_2020_CVPR} proposed a generative model that takes a 3D video slice as input and generates a single physical adversarial example. More recently, the method proposed by Yang et al.~\cite{yang2020finding} optimizes physical perturbations on a set of frame sequences and places them directly on the environment using a differentiable mapping of the perturbations in 2D space to the 3D space. However, both of these methods do not consider a target state for the agent to reach in the presence of physical adversarial examples. Boloor et al.~\cite{boloor2020attacking} showed a targeted attack on autonomous driving systems called a hijacking attack, where the agent takes a targeted path of actions pre-specified by the attacker. However, our approach differs by letting the attacker choose a final target state and using our attack algorithm to internally learn the path of actions to reach the target.


\vspace{-0.03in}
\section{Targeted Attack via Learned Visual Patterns of Physical Objects}
\vspace{-0.03in}
We formulate our task as attacking a deep RL system with the adversarial object to be continuously effective at misguiding the agent, while the agent is moving in the environment due to the dynamics of the agent. This is a key difference we claim from existing deep RL attacks that prior works consider. 
Such a requirement is necessary for the goal of manipulating the agent in a non-trivial way, leading to a guaranteed effective attack. Moreover, unlike perturbations in the state or actions spaces in existing attacks, we perturb a static rectangular area fixed to the environment. 
In the following section, we introduce the notation, problem statement, and technical details towards a solution.

\begin{figure*}[ht]
\centering
\includegraphics[width=50em]{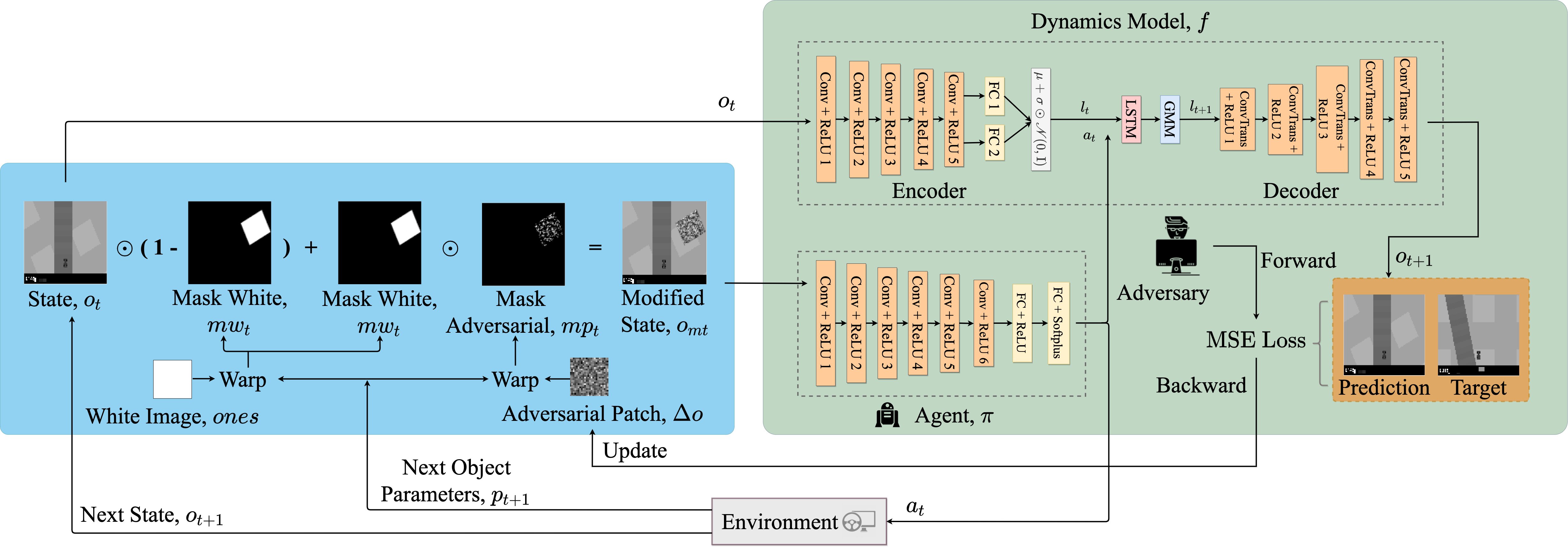}
\caption{Illustration of physical adversarial attack in OpenAI Gym's CarRacing-v0 environment. The blue panel shows the crafting of modified observation by adversary through planting and updating the physical object. Adversary creates a new dynamics model and it is assumed that pre-trained agent policy is known as shown in the green panel. The orange panel shows the optimization performed by adversary to perform physical adversarial attack by minimizing the loss between prediction from dynamics model and the predefined target observation.}
\vspace{-1em} 
\label{fig:attack}
\end{figure*}

\vspace{-0.05in}
\subsection{Notations and preliminaries}
\vspace{-0.03in}
\label{sec:notation}
Let $o_t \in [0,1]^{w \times h \times c}$ be a grey-scale image with width $w$, height $h$, and channel size $c$, that represents the state (scenes) of the underlying Markov decision process (MDP). In the experiment, $o_t$ is the stack of the last four top-down views of a driving scene, resembling a simplification of data obtained through LIDAR. We use $o_t$ as the most recent image of the stack. $a_t \in [0,1]^{n}$ is the action vector chosen by the agent at time $t$, and $n$ is the number of continuous actions to be determined. In the experiment, the actions include the normalized braking and acceleration rates and the change of steering angle. 

Let $\pi : [0,1]^{w \times h \times c} \to [0,1]^{n}$ be a deterministic policy learned on the MDP with $c$ equaling 1 to represent grayscale images, and let 
\begin{equation}
\vspace{-0.03in}
f: [0,1]^{w \times h \times c} \times [0,1]^{n} \to [0,1]^{w \times h \times c},    
\end{equation}
\vspace{-0.03in}
be the dynamics model of the environment that gives the next state $o_{t+1}$ when action $a_t$ is taken. We note that the agent, as a dynamical system, has its own state defined by normalized $\delta_t \in [0, 1]^{k}$, where $k$ is the number of properties. In the experiment, $\delta_t$ is represented by the position, velocity, and steering angle of the vehicle. We denote  the dynamics of the agent as:
\begin{equation}
\vspace{-0.03in}
    g:[0, 1]^{k} \times [0, 1]^{n} \to [0, 1]^{k}.
\end{equation}

We consider attacks in the form of a grey-scale image (perturbation) in a fixed rectangular area of the environment. This image, without transformation, is denoted by 
$\Delta o$, and its global coordinates by $\Phi$. To integrate this image into the 
scene ($o_t$), the following differentiable procedure is programmed:

\noindent(1) The relative position of the adversarial rectangle in the scene, denoted by $p_t$, is first calculated based on the agent dynamics, $g$, the object's global coordinates, $\Phi$, and a transformation function, $\psi$ as $p_t = \psi(\delta_t, \Phi)$, \text{where } $\delta_t = g(\delta_{t-1}, a_{t-1})$.

\noindent(2) Let $ones$ be a matrix of ones. A mask $mw_t \in \{0,1\}^{w \times h}$ is created based on $p_t$ and $ones$ using homography estimation, realized through the $warp$ function in Kornia~\cite{eriba2019kornia}. $mw_t$ only has $1$s within the rectangle. 

\noindent(3) A transformed adversarial image $mp_t \in [0,1]^{w \times h}$ is created based on $p_t$ and $\Delta o$, again using homography estimation. 

\noindent(4) Lastly, we integrate the adversarial image into the view:
\begin{equation}
\vspace{-0.03in}
o_{mt} = o_t \odot (1 - mw_t) + mw_t \odot mp_t,     
\end{equation}
\vspace{-0.03in}
where $\odot$ is the element-wise product. Although the homography estimation and warping procedure described above are similar to~\cite{yang2020finding}, our unique differentiable layer implementation allows solving through gradient-based methods rather than the local linearization approach presented in~\cite{yang2020finding}.

\vspace{-0.05in}
\subsection{Problem statement}
\vspace{-0.03in}
\label{sec:formulation}
Given the initial state $o_0$ (which contains duplicates of the initial scene), the initial agent state $\delta_0$, the pretrained policy $\pi$, the dynamical models $f(\cdot, \cdot)$ and $g(\cdot, \cdot)$, and the transformation function, $\psi(\delta, \Phi)$, we search for an image $\Delta o$, with $||\Delta o||_{\infty} \leq \epsilon$, that leads the agent to a specific target $o_{target}$ within the time window $[0,T]$. Formally:
\begin{equation}
\vspace{-0.03in}
\begin{aligned}
    \min_{||\Delta o||_{\infty} \leq \epsilon} \quad & \sum_{t=1}^{T} || o_t - o_{target}||_2^2.\\
    s.t. \quad & a_t = \pi(o_{mt}), \\
    & o_{mt} = o_t \odot (1 - mw_t) + mw_t \odot mp_t\\
    & mw_{t} = warp(ones, p_{t}), 
     mp_{t} = warp(\Delta o, p_{t}), \\
    & p_{t} = \psi(\delta_t, \Phi), 
     o_{t+1} = f(o_t, a_t), 
     \delta_{t+1} = g(\delta_t, a_t). \\
\end{aligned}
\label{eq:attack}
\end{equation}
\vspace{-0.03in}
The dependency of variables involved in this problem is visualized in Fig.~\ref{fig:attack}.
The loss function of Eq.~\eqref{eq:attack} accepts early convergence of the agent's state to the target. Notice that we use scenes without the adversarial perturbation in evaluating the loss, since the target state is specified before the attack problem is solved.   
The use of the learned dynamics model, agent's dynamics and a differentiable implementation of $warp$ together make this problem differentiable with respect to the perturbation $\Delta o$, allowing the problem to be solved using gradient-based methods.

\vspace{-0.05in}
\subsection{Learning dynamics of the environment}
\vspace{-0.03in}
Here we introduce the procedure for learning a differentiable dynamical model of the environment, which is an essential step to enable a gradient-based attack. We believe that addition of this dynamical model explicitly accounts for state evolution in the attack generation and also the plan of actions leading to target state. This makes our targeted attack more feasible and easier by letting the attacker specify a target state rather than how to reach that target state.

\subsubsection{Data collection}
We first collect data in the format of (state, action, next state) through multiple rollouts of the environment. Note that a successfully attacked rollout will encounter states different from those experienced through the benign policy, e.g., agent moving out of the highway. To collect a representative dataset, we perform rollouts using the pretrained policy with noise of variable strength $\tau$ added to the actions, i.e, $a_t = a_t + \mathcal{N}(0, 1) * \tau$ similar to~\cite{yang2020finding}.
The noisy actions help explore the environment, allowing the adversary to predict the environment dynamics correctly when approaching the target. The resultant dataset is denoted by $\mathcal{D} = \{(o_i, a_i, o_{i+1})\}_{i=1}^N$. We note that such data collection is achievable when launching a real-world attack, as long as the attacker can sample the state transitions towards the specified target by using a vehicle with dynamics similar to the attacked agent.

\subsubsection{Learning the environment dynamics}
Since the environment state contains rich information (e.g., time-variant track and surroundings), feed forward neural networks fail to generalize well on the dataset. Here we follow Ha et al.~\cite{ha2018recurrent} to construct a dynamical model using a variational autoencoder (VAE) and a mixture-density recurrent neural network (MD-RNN), denoted by $\hat{f}(\cdot, \cdot; w)$, which takes in the environment state and action, and predicts the next environment state. $w$ are trainable parameters. As in \cite{ha2018recurrent}, we use the same combination of mean square error and Kullback–Leibler divergence as the loss for training the VAE, and the Gaussian mixture loss for training the MD-RNN.

\vspace{-0.05in}
\subsection{Optimization details}
\vspace{-0.03in}
We use Alg.~\ref{alg:optim} to solve the attack problem (Eq.~\eqref{eq:attack}). 
During each iteration, we obtain the state containing the adversarial image $o_{mt}$ as described in Eq.~\ref{eq:attack} by computing $mp_t$ and $mw_t$. To respect the observation limits seen by the agent, we clip $o_{mt}$ between $0$ and $1$ so that a valid image is yielded. The agent then performs an action on $o_{mt}$ to get $a_t$. Using the dynamics model, $f$, future prediction $o_{t+1}^{\dagger}$ is obtained to compute the loss. Finally, we backpropagate the sum of losses within the time window $[0,T]$ in order to update perturbation $\Delta o$.

\begin{algorithm}[t]
\small
\textbf{Input:} Number of Iterations, $I$, environment $env$, Attack length, $T$,
pretrained policy $\pi$, dynamics model, $f$, target state $o_{target}$ \\
\textbf{Output:} learned physical perturbation example, $\Delta o$ \\
$i \gets 0$, seed $\gets$ random seed \\
$\Delta o \gets \mathcal{N}(0, 1)$ \\
\While{$i < I$}{
  total\_loss $ \gets 0$, $t \gets 0$ \\
  $env$.seed(seed) \\
  $o_t = env$.reset() \\
  $\delta_t \gets $ initial agent state\\
  \While{$t < T$} {
    $p_t = \psi(\delta_t, \Phi)$\\
    $mw_t, mp_t = warp(ones, p_t), warp(\Delta o, p_t)$\\
    $o_{mt} = o_t \odot (1 - mw_t) + mw_t \odot mp_t$ \\
    clip $o_{mt}$ between $[0, 1]$ \\
    $a_t = \pi(o_{mt})$ \\
    $o_{t+1}^{\dagger} = \hat{f}(o_t, a_t)$ \\
    $\delta_{t+1} = g(\delta_t, a_t)$ \\
    total\_loss += $d(o_{t+1}^{\dagger}, o_{target})$ \\
    $o_{t+1} \gets env$.step($a_t$) \\
    $t \gets t + 1$
  }
  backpropagate total\_loss to update $\Delta o$ \\
  clip $\Delta o$ between $[-\epsilon, \epsilon]$ \\
  $i \gets i + 1$
}
\textbf{Return $\Delta o$}
\caption{Optimization for Targeted Physical Adversarial attack}
\label{alg:optim}
\end{algorithm}



\newcommand{\sconestart}{\includegraphics[width=7em]{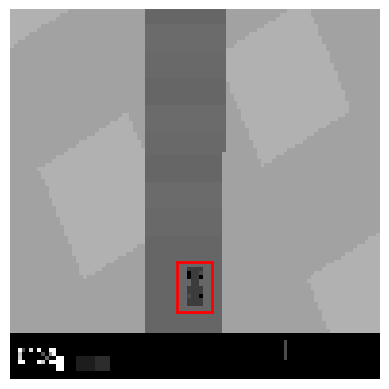}}
\newcommand{\sconetarget}{\includegraphics[width=7em]{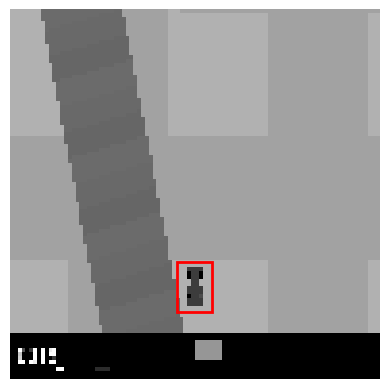}}
\newcommand{\sconestartobj}{\includegraphics[width=7em]{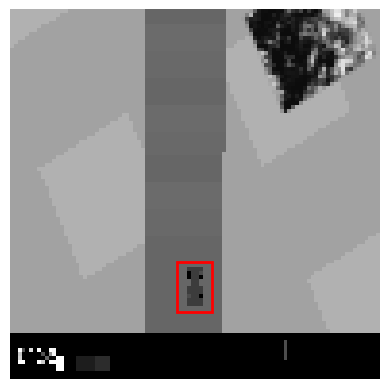}}
\newcommand{\sconeendnoobj}{\includegraphics[width=7em]{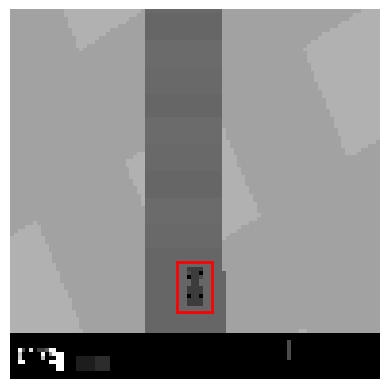}}
\newcommand{\sconeendobj}{\includegraphics[width=7em]{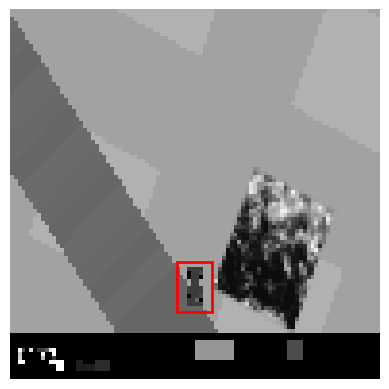}}

\newcommand{\sctwostart}{\includegraphics[width=7em]{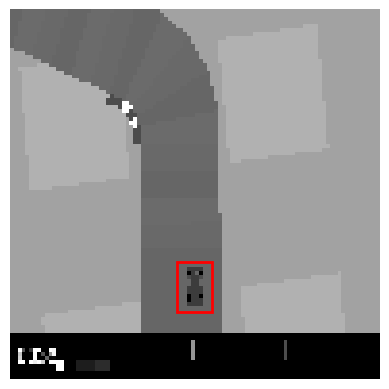}}
\newcommand{\sctwotarget}{\includegraphics[width=7em]{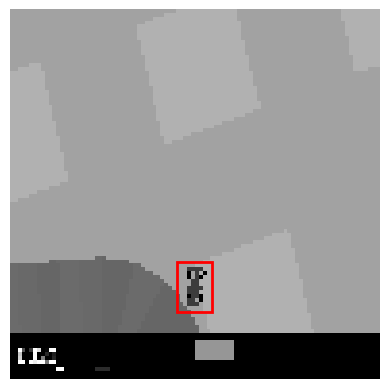}}
\newcommand{\sctwostartobj}{\includegraphics[width=7em]{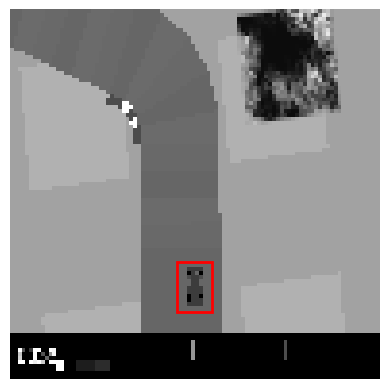}}
\newcommand{\sctwoendnoobj}{\includegraphics[width=7em]{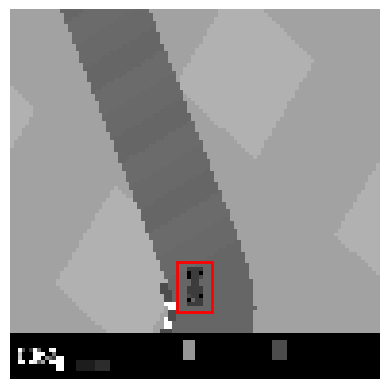}}
\newcommand{\sctwoendobj}{\includegraphics[width=7em]{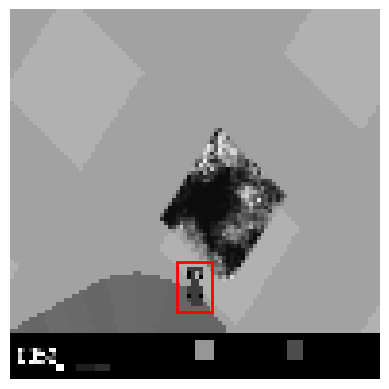}}

\newcommand{\scthreestart}{\includegraphics[width=7em]{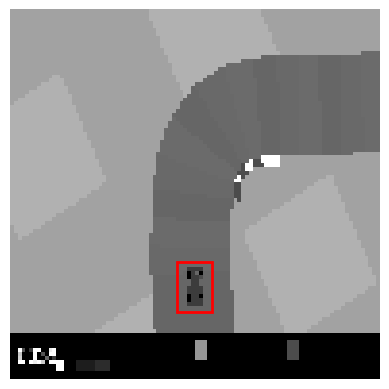}}
\newcommand{\scthreetarget}{\includegraphics[width=7em]{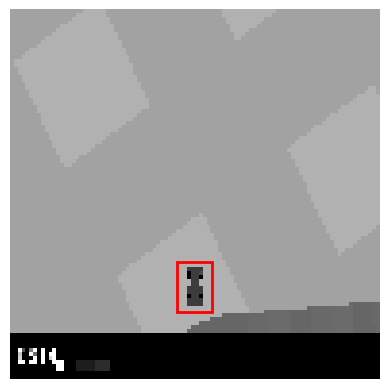}}
\newcommand{\scthreestartobj}{\includegraphics[width=7em]{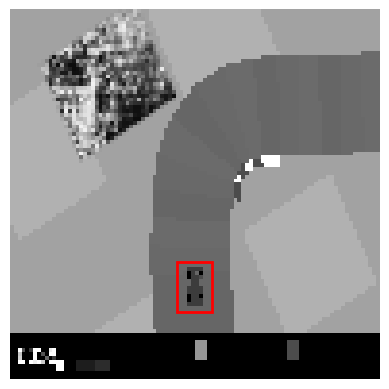}}
\newcommand{\scthreeendnoobj}{\includegraphics[width=7em]{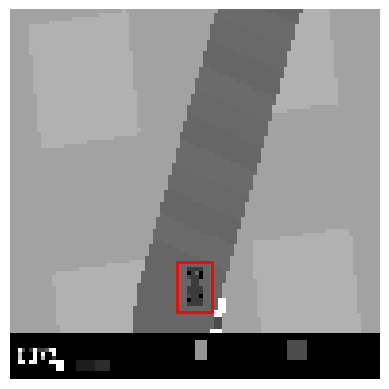}}
\newcommand{\scthreeendobj}{\includegraphics[width=7em]{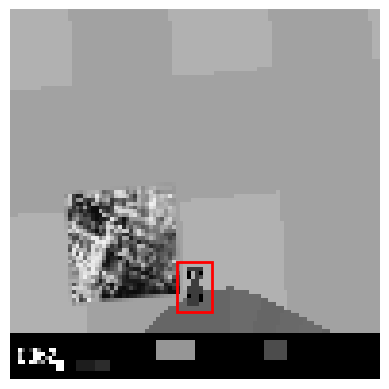}}

\newcommand{\sconerandom}{\includegraphics[width=7em]{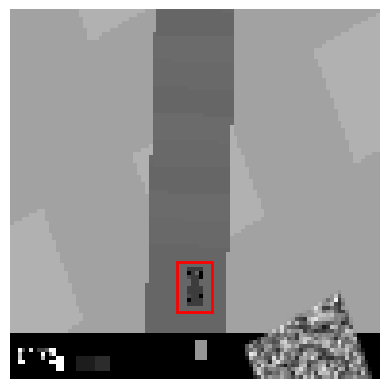}}
\newcommand{\sctworandom}{\includegraphics[width=7em]{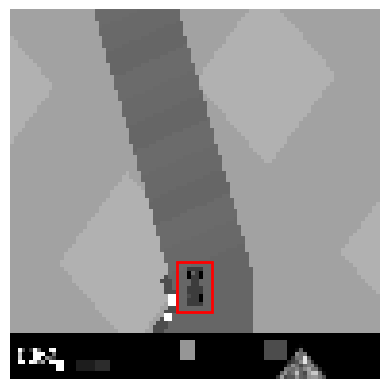}}
\newcommand{\scthreerandom}{\includegraphics[width=7em]{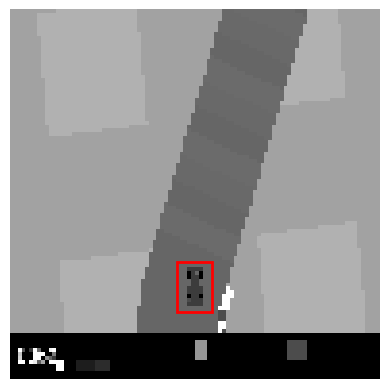}}

\begingroup
\setlength{\tabcolsep}{10pt} 
\renewcommand{\arraystretch}{2} 
\newcolumntype{C}{>{\centering\arraybackslash}m{6em}}
\begin{table*}
\centering
\caption{Targeted and random attacks in three scenarios. Agent in red boxes. See supplementary video for details.}
\begin{tabular}{l|CC|CCC|C}
  \hline
    Scenarios & ($t=0$) No attack & Optimal attack
    & ($t=T$) No attack & Optimal attack & Random attack & Target state
    \\
    \hline
    Straight & \sconestart & \sconestartobj & \sconeendnoobj & \sconeendobj & \sconerandom & \sconetarget \\
    Left turn & \sctwostart & \sctwostartobj & \sctwoendnoobj & \sctwoendobj & \sctworandom & \sctwotarget\\
    Right turn & \scthreestart & \scthreestartobj & \scthreeendnoobj & \scthreeendobj & \scthreerandom  & \scthreetarget\\
    \hline
\end{tabular}
\label{tab:attack_ill}
\end{table*}

\setlength{\textfloatsep}{8pt plus 1.0pt minus 2.0pt}
\begin{figure*}[t]
\centering
\begin{subfigure}{.3\textwidth}
  \centering
  \includegraphics[width=13em]{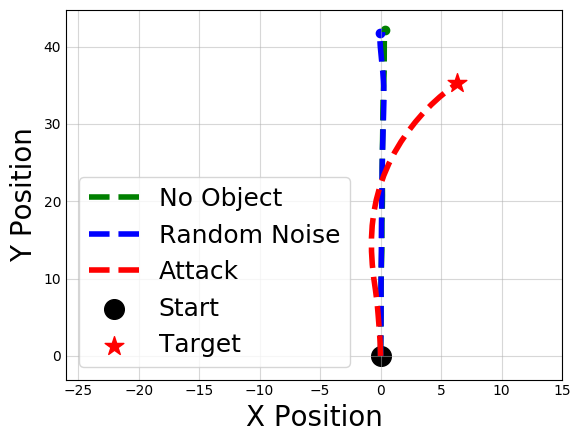}
  \caption{Straight track}
  \label{fig:deviation_sc1}
\end{subfigure}%
\begin{subfigure}{.3\textwidth}
  \centering
  \includegraphics[width=13em]{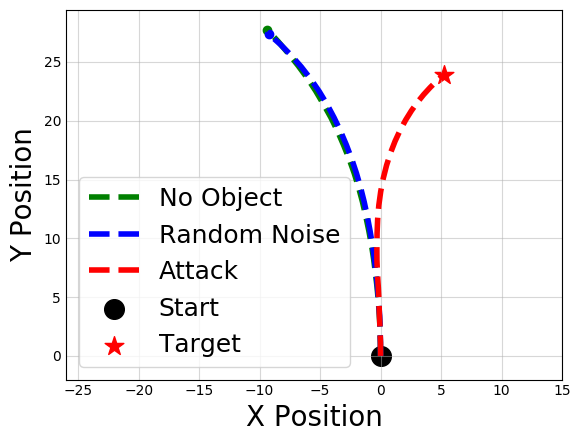}
  \caption{Left turn track}
  \label{fig:deviation_sc2}
\end{subfigure}
\begin{subfigure}{.3\textwidth}
  \centering
  \includegraphics[width=13em]{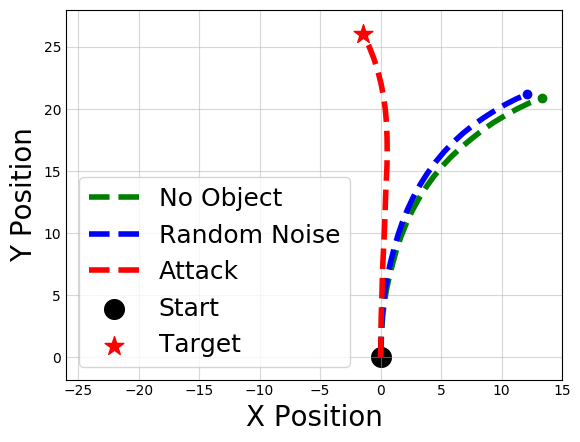}
  \caption{Right turn track}
  \label{fig:deviation_sc3}
\end{subfigure}
\caption{Trajectories in the three scenarios with no attack, random attack, and optimized attacks.}
\vspace{-1em} 
\label{fig:deviation}
\end{figure*}

\vspace{-0.05in}
\section{Experiments}
\vspace{-0.05in}
We use the CarRacing-v0 environment~\cite{carracing_v0} in OpenAI Gym to demonstrate the existence of adversarial objects that misguide an otherwise benign deep RL agent. We used a model-free Actor-Critic algorithm~\cite{pmlr-v48-mniha16} to obtain the pretrained policy $\pi$. The policy is trained with a batch size of 128 and $10^5$ episodes. 

For the dynamical model $\hat{f}$ of the environment, the VAE is trained for $10^3$ epochs using the Adam optimizer~\cite{kingma2014adam}. We set the batch size to 32 and learning rate to $0.001$ with decreasing learning rate based on plateau and early stopping. For the MD-RNN, we train for $10^3$ epochs using the same optimizer. We set the batch size to 16, the number of Gaussian models to 5, and the learning rate to the same value as the training of VAE.

For the attack, we set the time span $T$ to 25 and the adversarial bound to $\epsilon = 0.9$. An ablation study will be done on these hyperparameters in Sec.~\ref{sec:ablation}. We use the same optimizer as before, and set the learning rate to $0.005$ for $I = 10^3$ iterations. We set the adversarial area to be 25 pixels wide and 30 pixels tall. 

\subsection{Baselines}
To the best of our knowledge, there have been few results on targeted physical attacks on deep RL agents. Although the work of Yang et al.~\cite{yang2020finding} is similar to ours, we believe that their experiment setting is very different from ours and we thus did not use it as a baseline. Therefore, we use a baseline where $\Delta o$ is drawn uniformly in $[0,1]^{25 \times 30}$. By comparing agent state trajectories in the presence of random and optimized $\Delta o$, we will show that the proposed attack is more effective than random perturbations.

\subsection{Evaluation metrics}
We introduce two metrics to evaluate the effectiveness of an attack: \textit{actions error} and \textit{percentage change of value}.
The former is defined as the mean square error between the attacked and benign action values over $T$ timesteps derived from rollouts with and without the adversarial object, respectively.
The latter is the percentage change of value from the benign to the attacked rollout, where the value of a policy is the sum of rewards over $[0,T]$.

\section{Experiments and Discussion}
We evaluate these metrics on three driving scenarios, and compare the trajectories of the agent with and without the attack.
Further, we conduct experiments to evaluate the robustness of our attack with respect to different locations of object.
Finally, we compare the effectiveness of the attack with varying time span ($T$) and adversarial bound ($\epsilon$) based on the evaluation metrics.

\subsection{Attack scenarios}
We consider three driving scenarios where the agent with the benign policy will go straight, left, and right, respectively. In each of the scenarios, the object is placed at a fixed location in the environment so that it is observable by the agent throughout the attack. We specify the target states as the images shown in Table~\ref{tab:attack_ill}.



\subsection{Comparison with the baseline}
We compare the trajectories of the agent under the benign policy, the proposed attack, and the random attacks, with trajectory visualizations in Fig.~\ref{fig:deviation} and final states in Tab.~\ref{tab:attack_ill}. For the random attack, we conducted 10 independent simulations for each scenario to derive the mean trajectories. The standard deviations in all three scenarios are negligible. $X$ and $Y$ axes in the figure represent the global coordinates. 

Results show that while our approach successfully misguides the agent in all scenarios, the agent is not affected as much by the random attacks.
Specifically, in scenario 1, the agent goes straight with and without the presence of a random attack. In the presence of the proposed attack, however, the agent deviates from the benign trajectory to reach the target state.
The same happens for scenarios 2 and 3.
It is worth noting that by comparing Tab.~\ref{tab:attack_ill} and Fig.~\ref{fig:deviation}, we see that the agent reaches the target location but does not perfectly match the target orientation. For instance, in the straight track scenario, we can see that the optimal attack after $t=T$ time steps forces the car to turn right, partially going off the road, but in the target state, the car is completely off the road on the right. Further exploration of the attack objective may potentially improve the matching of the orientation.
Lastly, Table~\ref{tab:loss_action_reward} quantifies the comparison through the evaluation metrics. The proposed attack outperforms the random ones on both metrics. 

\begingroup
\setlength{\tabcolsep}{6pt} 
\renewcommand{\arraystretch}{1.5} 
\begin{table}[]
    \centering
    \caption{Comparison with random noise baseline in terms of evaluation metrics.}
    \begin{tabular}{l|ccc}
    \hline
    \textbf{\thead{Scenarios}} & \textbf{\thead{Actions Error}} & \textbf{\thead{Change of value (\%)}}\\
    \hline
    Straight + Random & 0.064 & 0 \\
    Left turn + Random & 0.069 & 0 \\
    Right turn + Random & 0.046 & -10.72 \\
    \hline
    Straight + Proposed & \textbf{0.126} & \textbf{-17.70} \\
    Left turn + Proposed & \textbf{0.138} & \textbf{-32.26}\\
    Right turn + Proposed & \textbf{0.062} & \textbf{-32.15}\\
    \hline
    \end{tabular}
    \label{tab:loss_action_reward}
\end{table}

\begin{figure}[ht!]
\centering
\includegraphics[width=15em]{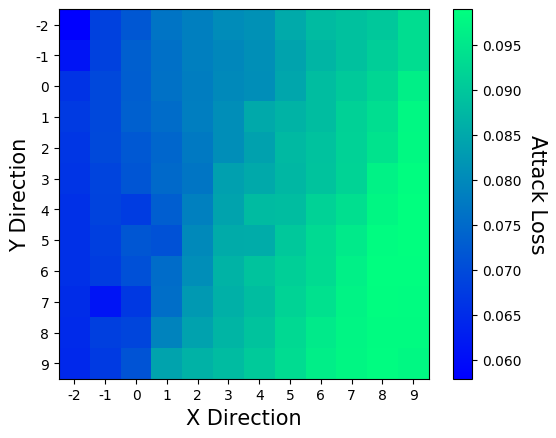}
\caption{Attack robustness heatmap on position of physical object in scenario 1}
\label{fig:robust}
\end{figure}

\subsection{Robustness to translation}
We study the robustness of our attack with respect to different global coordinates of the attack object ($\Phi$) placed in the environment. Specifically, we changed the position of adversarial object iteratively in $x$ and $y$ directions during test time with the same learned adversarial pattern. The experiment is performed on the straight track scenario, with fixed dynamical models. In Fig.~\ref{fig:robust}, the $(x,y)$ coordinates represent position of the attack object relative to the actual object position, $(0, 0)$ in the original attack (i.e., the one used in the experiments for Tab.~\ref{tab:attack_ill}), and the heat map of Fig~\ref{fig:robust} represents the attack loss of Eq.~\eqref{eq:attack}, when the object is moved accordingly.
Therefore the blue region represents more successful attacks since the attack loss is lower, whereas the green region represents relatively unsuccessful attacks as attack loss is higher.
Note that the range of the figure is bounded by the constraints that the object cannot be on the track and cannot be out of the scene.
From this test, if the object is moved towards the track ($-X$ direction in the figure), the attack will still be effective or even better. 
On the other hand, if the object is moved away from the track and partially out of the scene, the attack becomes less effective, which is reasonable since the agent will have only partial observation of the attack. Further investigating formulations of robust attacks will be valuable.

\subsection{Adversarial bounds and Attack length} \label{sec:ablation}
\begingroup
\setlength{\tabcolsep}{7pt} 
\renewcommand{\arraystretch}{1.6} 
\begin{table}[ht]
\centering
\caption{Adversarial Bounds \textit{vs} Attack Length}
\begin{tabular}{|c|c|c|c|c|c|c|c|c|c|}
\hline
\multirow{3}{*}{\makebox[4em]{\textbf{\thead{Adversarial \\Bound $\epsilon$}}}} & \multicolumn{6}{c}{\textbf{Attack Length $T$}} \vline \\
\cline{2-7}
& \multicolumn{2}{c}{\textbf{$T$ = 15}} \vline
& \multicolumn{2}{c}{\textbf{$T$ = 25}} \vline 
& \multicolumn{2}{c}{\textbf{$T$ = 30}} \vline \\
\cline{2-7}
& \makebox[1em]{\textbf{\thead{Attack \\Loss}}} 
& \makebox[1em]{\textbf{\thead{Actions \\Error}}}  
& \makebox[1em]{\textbf{\thead{Attack \\Loss}}}     
& \makebox[1em]{\textbf{\thead{Actions \\Error}}}       
& \makebox[1em]{\textbf{\thead{Attack \\Loss}}}
& \makebox[1em]{\textbf{\thead{Actions \\Error}}} \\
\hline
0.1  &   0.091 & 0.064   &  0.090 & 0.064   &   0.088 & 0.063 
\\ \hline
0.3 &   0.088 & 0.078    &   0.087 & 0.069   &   0.085 & 0.066
\\ \hline
0.5  &   0.086 & 0.113  &   0.077 & 0.107  & 0.083 & 0.070
\\ \hline
0.9  &   0.081 & 0.125   &   \textbf{0.076} & \textbf{0.126} & 0.078 & 0.093
\\ \hline
\end{tabular}
\label{tab:attack_vs_bounds}
\end{table}

Here we perform an ablation study on the attack strength ($\epsilon$) and attack time span ($T$), by enumerating $\epsilon \in \{0.1, 0.3, 0.5, 0.9\}$ and $T \in \{15, 25, 30\}$. The results in terms of the optimal loss of Eq.~\eqref{eq:attack}, and the actions error are summarized in Table~\ref{tab:attack_vs_bounds}. 
The experiments are performed on the straight track scenario, with fixed dynamical models. 
From the results, it is evident that larger $\epsilon$ improves the effectiveness of the attacks. Additionally, as we enlarge the time window, the actions error decreases in nearly all cases. Based on our experiments, we believe that if $T$ is smaller, then the attack has a smaller action space to plan on, causing it to alter the actions more aggressively than a bigger $T$. However, the attack loss increases from $T = 25$ to $T = 30$. By examining the results, we found that this is primarily because the attack object moves out of the scene between $T=25$ and $T=30$. As a result of the limited observation of the attack object by the agent, the optimizer struggles to find a way to keep the agent close to the target state, thus the increased loss. This implies that the time window is coupled with the choice of the target state, and its careful selection is important for succeeding in the attack.

\vspace{-0.03in}
\section{Conclusion} 
\vspace{-0.03in}
\label{sec:conclusion}
Even though autonomous driving agents have been increasingly using deep RL techniques, it is possible that they can be fooled by simply placing an adversarial object in the environment. While previous studies in this domain focused on untargeted attacks without long-term effects, this paper studies the \textit{existence} of static adversarial objects that can continuously misguide a deep RL agent towards a target state within a time window. Using a standard simulator and a pretrained policy, we developed an algorithm that searches for such attacks and showed their existence empirically. For effective search of the attacks, we utilize differentiable dynamical models of the environment, which can be learned through experience collected by the attacker.
Our approach has a limitation that the full policy must be known to the attacker (white-box). Additionally, the attack highly depends on the size, position, and pattern of the object. More improvements on these areas are necessary to better understand the practicality of the threat model.
Future work will study the existence of robust physical attacks in more complex environments, e.g., with the presence of other agents and with visual or 3D observations.
By demonstrating the existence of a new type of attack more practical than digital perturbations, we hope this study can motivate more rigorous research towards robust and safe AI methods for autonomous driving.

\section{Acknowledgments.}
This material is based upon work supported by the National Science Foundation under Grant No. \#1925403, \#2038666 and \#2101052. The authors acknowledge support from Amazon AWS Machine Learning Research Award (MLRA), and the Institute of Automated Mobility (IAM), Arizona Commerce Authority. 
Any opinions, findings, and conclusions or recommendations expressed in this material are those of the author(s) and do not necessarily reflect the views of the funding entities.

\newpage
\bibliographystyle{IEEEtran}
\bibliography{IEEEabrv,references}

\end{document}